\documentclass[10pt,twocolumn,letterpaper]{article}

\usepackage{iccv}
\usepackage{times}
\usepackage{epsfig}
\usepackage{graphicx}
\usepackage{amsmath}
\usepackage{amssymb}
\usepackage{times}
\usepackage{epsfig}
\usepackage{graphicx}
\usepackage{amsmath}
\usepackage{amssymb}
\usepackage{indentfirst} 
\usepackage[linesnumbered,ruled,vlined]{algorithm2e}
\usepackage{graphicx}
\usepackage{amsmath}
\usepackage{amssymb}
\usepackage{booktabs}
\usepackage[title]{appendix}
\usepackage{cuted}
\usepackage{subfigure}
\usepackage{marvosym}

\usepackage{multirow}
\usepackage{diagbox}
\usepackage{colortbl}  
\usepackage{array}   
\usepackage{ulem}

\usepackage{cite}

\usepackage[breaklinks=true,bookmarks=false]{hyperref}

\iccvfinalcopy 

\def\iccvPaperID{****} 

\ificcvfinal\fi

\begin{document}
\title{Edge-aware Plug-and-play Scheme for Semantic Segmentation}  

\author{Jianye Yi\thanks{These authors contributed to the work equally and should be regarded as co-first authors.}, Xiaopin Zhong\footnotemark[1],
Weixiang Liu\textsuperscript{\Letter{}}, \\
Wenxuan Zhu, Zongze Wu, Yuanlong Deng\\
Lab. of Machine Vision and Inspection, College of Mechatronics and Control Engineering, \\
Shenzhen University,
\#3688 Nanhai Ave, Shenzhen, PR China\\
{\tt\small 2110296017@email.szu.edu.cn, xzhong@szu.edu.cn, \textsuperscript{\Letter{}}wxliu@szu.edu.cn,}\\ {\tt\small 2110296009@email.szu.edu.cn,zzwu@szu.edu.cn, dengyl@szu.edu.cn}}

\date{}
\maketitle
\thispagestyle{empty}
\setlength{\parindent}{2em}
\begin{abstract}
Semantic segmentation is a classic and fundamental computer vision problem dedicated to assigning each pixel with its corresponding class. 
Some recent methods introduce edge-based information for improving the segmentation performance. 
However these methods are specific and limited to certain network architectures, and they can not be transferred to other models or tasks. 
Therefore, we propose an abstract and universal edge supervision method called Edge-aware Plug-and-play Scheme (EPS), which can be easily and quickly applied to any semantic segmentation models. The core is edge-width/thickness preserving guided for semantic segmentation.
The EPS first extracts the Edge Ground Truth (Edge GT) with a predefined edge thickness from the training data; and then for any network architecture, it directly copies the decoder head for the auxiliary task with the Edge GT supervision.
To ensure the edge thickness preserving consistantly, we design a new boundary-based loss, called Polar Hausdorff (PH) Loss, for the auxiliary supervision.
We verify the effectiveness of our EPS on the Cityscapes dataset using 22 models. 
The experimental results indicate that the proposed method can be seamlessly integrated into any state-of-the-art (SOTA) models with zero modification, resulting in promising enhancement of the segmentation performance.

\end{abstract}

\section{Introduction}
Semantic segmentation aims to achieve pixel-level classification by providing dense predictions for each pixel.
With the rapid development of convolutional neural networks and the application of Transformer\cite{vaswani2017attention} in the field of computer vision, a series of deep learning-based semantic segmentation models have emerged, such as CNN-based FCN\cite{long2015fully}, DeepLab\cite{chen2017deeplab}, PSPNet\cite{zhao2017pyramid}, CGNet\cite{wu2020cgnet}, and ViT-based Segmenter\cite{strudel2021segmenter} and SegFormer\cite{xie2021segformer}, etc.
Researchers are always striving to propose new network structures to improve the performance of semantic segmentation. 
They usually approach the problem from the perspective of model design, resulting in specific and unique network structures. 
However, this approach may lead to overfitting during training and may not be easily applicable to various applications.
Therefore, we believe that it is more effective to design a general and abstract scheme that is applicable to any model, rather than solely relying on model design.

Currently, in supervised semantic segmentation tasks, most researchers directly use the original annotated data for supervision. 
However, a minority of researchers have explored other features of the original data for more effective supervision, such as adding edge supervision. 
Edge detection is a task of extracting image edges\cite{marr1980theory}. 
In recent years, related works\cite{zheng2019elkppnet,zhang2019net,li2020improving,hatamizadeh2020edge,chen2020semeda,chen2016semantic} have embedded the results of edge detection into semantic segmentation and confirmed that edge supervision can effectively improve the accuracy of segmentation models.
Due to the locality of CNN's inductive bias, it is necessary to perform pooling on the feature maps to increase the receptive field (RF), which leads to blurring and uncertainty of segmentation boundaries\cite{zhou2014object,luo2016understanding}, ultimately limiting segmentation accuracy. 
Although ViT-based segmentation models have global RF, there is currently no related work on adding edge supervision to ViT-based segmentation models. However, this does not mean that edge supervision is not important for ViT-based segmentation networks.
Whether from a spatial-geometric perspective, dividing the objects of semantic segmentation into edges and bodies, or from a frequency domain perspective, dividing them into high-frequency and low-frequency information, these classifications are based on human experience. 
This partly explains why adding edge supervision as prior knowledge can improve model accuracy.

To leverage edge information as prior knowledge to improve segmentation performance, researchers have incorporated an edge supervision task into the network. 
Li et al.\cite{li2020improving} decoupled the edges and bodies of the targets and supervised them separately, and then fused the body feature and the residual edge feature.
However, this approach requires specially designed decouplers and fusers, which are not easily transferable to other segmentation models. 
Zhang et al.\cite{zhang2019net} designed an auxiliary decoder that utilizes the edge features extracted from the first two layers of the CNN's multi-scale features for edge supervision. 
However, this method is only suitable for CNN networks and is not applicable to ViT-based segmentation networks, as ViT does not have multi-scale features. 
Chen et al.\cite{chen2020semeda} proposed a SEMEDA framework with a segmentic edge detection network to extract edges from segmentation results for supervision. 
However, this structure is specific and may not be effective in other segmentation networks. 
To the best of our knowledge, all existing techniques that integrate edge supervision tasks rely on network architectures tailored for this purpose and are deficient in rapid transfer and universal applicability attributes.
In addition, they use distribution-based cross-entropy loss, which does not have spatial-geometric characteristics and is therefore unsuitable for edge images with spatial-geometric features.

We propose an Edge-aware Plug-and-play Scheme (EPS) to address the issue of edge supervision modules lacking the characteristics of easy plug-and-play and general applicability in semantic segmentation. 
This scheme exhibits universal applicability across various networks and is solely utilized during the training phase, ensuring that the model size and inference speed remain unaltered during testing.
Additionally, we propose a Polar Hausdorff (PH) Loss, a simplified version of the Hausdorff distance (HD) Loss represented in polar coordinates, to better utilize edge information. 
With the given edge thickness $d_e$, EPS can calculate the kernel size to generate Edge GT, and the optimization target of PH Loss is to constrain the edge thickness in the edge segmentation result to approach $d_e$. 
We conduct experiments on the Cityscapes dataset with 22 models in the MMSegmentation framework to demonstrate the effectiveness of EPS.
Our contributions are summarized as follows:
\begin{itemize}
    \item We introduce a novel scheme called EPS that offers effective edge supervision to any semantic segmentation model, regardless of whether it is based on CNN or ViT.
    \item In order to leverage edge information, we propose a novel boundary-based loss function, PH Loss, which restricts the thickness of edges and improves the accuracy of edge supervision signals.
    \item Our experiments show that EPS can be seamlessly integrated into the existing SOTA without any modification, leading to improved model accuracy.
\end{itemize}

\section{Related Work}

\subsection{Edge-supervised Segmentation}
\begin{figure}[h]
    \centering
    \includegraphics[width=8cm]{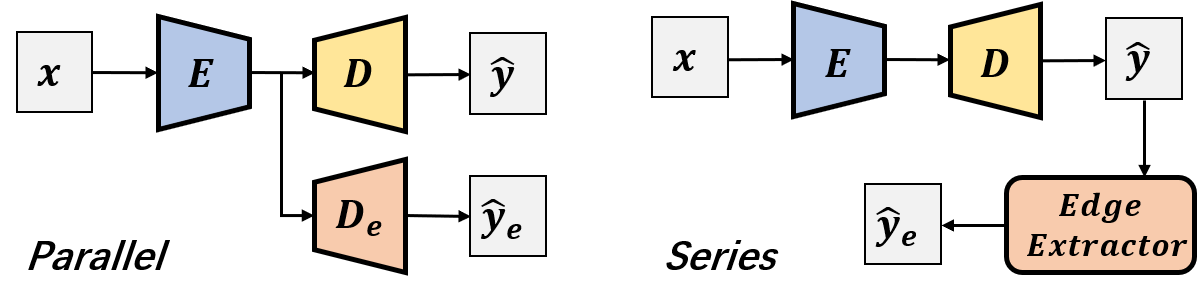}
    \caption{The left-hand side shows a parallel framework with edge supervision, while the right-hand side shows a serial framework with edge supervision. $x$ denotes the input image, $\hat{y}$ represents the GT, and $\hat{y}_e$ is the Edge GT. $E$ and $D$ denote Encoder and Decoder, respectively, while $D_e$ is the Edge Decoder.}
    \label{fig:edge_supervised}
\end{figure}

In semantic segmentation tasks, the higher the segmentation accuracy of the target, the more accurate the segmentation of its edge parts tends to be, and vice versa.
Therefore, effectively incorporating edge supervision in the segmentation model can improve the segmentation accuracy of the network.
Currently, there are two main approaches to incorporating edge supervision in segmentation models: parallel supervision and 
series supervision (as shown in Figure \ref{fig:edge_supervised}).

The parallel supervision usually adds an auxiliary head outside the backbone, which takes the original input image $x$ as its input and uses the edge label $\hat{y}_e$ for supervision to improve the segmentation accuracy of the decoder head on the backbone.
For example, EG-CNN \cite{hatamizadeh2020edge} uses edge gated layers to reconstruct the edges of the target, while ET-Net \cite{zhang2019net} uses the feature extraction ability of the first two layers of CNN to extract low-level features map for edge supervision.
However, these methods are not suitable for ViT-based models that do not have multi-scale features. Chen et al. \cite{chen2016semantic} combines edge detection tasks with segmentation tasks, and through a fusion network, fuses the results of edge detection and segmentation to improve the segmentation accuracy.
Moreover, Li et al. \cite{li2020improving} first extracts the edges from the segmentation results, and then uses edge and body supervision to separately segment the edges and bodies, and finally designs a fusion model to merge the results of both parts.
However, the latter two methods require an additional fusion network to use edge supervision to improve segmentation accuracy. Therefore, the approach of extracting edges and segmentation results separately and then merging them is cumbersome.

The series supervision involves the attachment of an auxiliary edge detector to a segmentation network, with input provided by $\hat{y}$, the predicted output from the segmentation network. 
Through supervision using edge labels $\hat{y}_e$, the information extraction capabilities for edges in the backbone and decoder are influenced, resulting in improved segmentation accuracy.
Zheng et al. proposed KLPPNet \cite{zheng2019elkppnet} that extracts edges through traditional algorithmic methods directly from the segmentation results and calculates its loss against edge labels $\hat{y}_e$.
SEMEDA \cite{chen2020semeda}, on the other hand, introduced a segmentic edge detection network in series framework connected to the decoder head to exploit edge supervision information.
However, such a series approach increases the architecture's depth, which may trigger adverse effects on back-propagation outcomes during edge supervision and potentially lower the primary network's edge information extraction capacity.

Most of the previous works above are designed with specific edge supervision modules for certain networks, whether in parallel or series supervision. 
However, such specific modules are difficult to transfer into other segmentation networks and their effectiveness cannot be guaranteed.

\subsection{Boundary-based Loss}
\begin{figure*}[h]
    \centering
    \includegraphics[width=14cm]{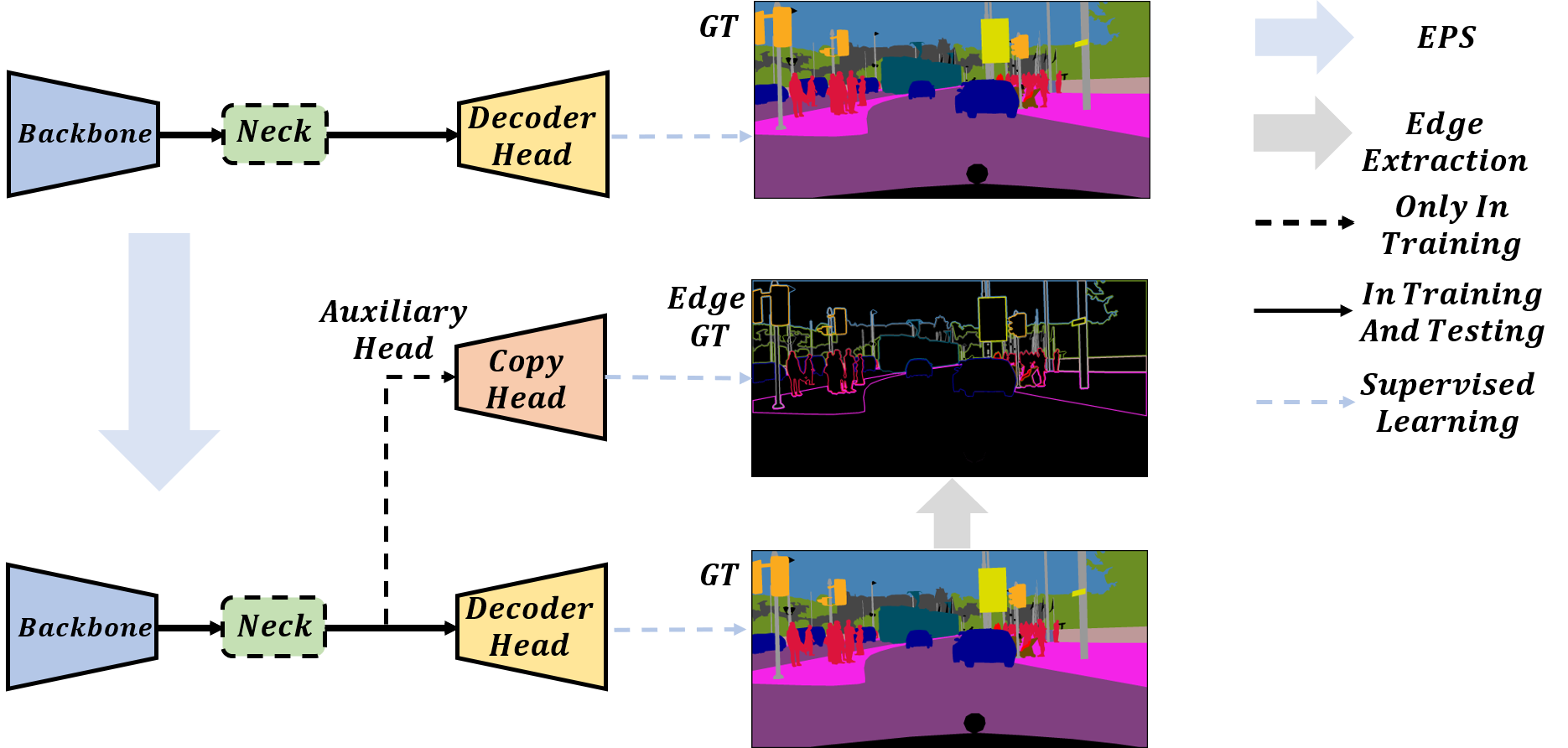}
    \caption{For a semantic segmentation model with only one decoder head, the EPS involves creating a new auxiliary head by completely copying the original decoder head, without changing any of its structure. Then, the Edge GT obtained from GT by edge extraction is used for edge supervision. For a semantic segmentation model that already has an auxiliary head, we directly replace the GT with Edge GT to perform edge supervision on its auxiliary head without any other operations.
    }
    \label{fig:CopyHead_only_decoder_head}
\end{figure*}

In semantic segmentation, the loss function plays a crucial role as it can significantly affect network learning efficiency.
Existing segmentation losses have been classified into four categories by Ma et al. \cite{ref_Ma_Loss} and Jadon et al. \cite{ref_Jadon_survey}: distribution-based losses, region-based losses, compound losses, and boundary-based losses.
Different types of loss functions have different optimization objectives and focuses, where distribution-based losses aim to improve overall classification accuracy, region-based losses aim to increase the overlap between predicted results and true labels, and compound losses combine the strengths of both types. 
The boundary-based loss, on the other hand, approaches it from a spatial geometry perspective by using the distance between the predicted and ground truth labels' boundaries to construct the loss function.
This study is centered on boundary-based loss functions, wherein some of the premier functions within this classification encompass the Boundary (BD) Loss \cite{kervadec2019boundary} and the Hausdorff distance (HD) Loss \cite{karimi2019reducing}.

\textbf{BD Loss} measures the quality of boundary prediction by calculating the distance between each pixel and the nearest boundary pixel in GT. The specific calculation formula can be defined as  \cite{kervadec2019boundary}:
\begin{equation}
 \mathcal\phi_G(q) =\left\{
\begin{array}{rcl}
-D_G(q),       &      & {q \in G}\\
D_G(q),       &      & {q \not\in G}
\end{array} \right. ,     
\label{equ:phi_G(q)}
\end{equation}
\begin{equation}
\begin{array}{rcl}
    \mathcal L_{BD}=\int_{\Omega}\phi_G(q)S_\theta(q)dq,
\end{array}
    \label{equ:L_BD}
\end{equation}
where, $\Omega$ refers to the entire image area, $q \in \Omega$ is any pixel point on the image, $G \subseteq \Omega$ is the region where the GT exists, with binary pixel values of $\{0,1\}$, $S_{\theta} \subseteq \Omega$ is the predicted labeling area, with pixel values of $(0,1)$, and $D_G(q)$ is the distance between the pixel point $q$ and the nearest pixel points on the boundary of the region $G$.

\textbf{HD Loss} measures the accuracy of boundary prediction by computing the Hausdorff distance between predicted and GT boundaries. Its formula can be expressed as:
\begin{equation}
\begin{array}{rcl}
     d_{AB}=max_j(min_i(d(a_i,b_j))),
\end{array}
    \label{equ:d_NM}
\end{equation}
\begin{equation}
\begin{array}{rcl}
     d_{BA}=max_j(min_i(d(a_j,b_i))),
\end{array}
    \label{equ:d_NM}
\end{equation}
\begin{equation}
\begin{array}{rcl}
    \mathcal L_{HD} = max(d_{AB}, d_{BA})
\end{array}
    \label{equ:d_NM}
\end{equation}
where, $A$ is the total number of pixels on the predicted boundary, $a_i$ represents a pixel on it, $B$ is the total number of pixels on the GT boundary, $b_j$ represents a pixel on it, and $d(a_i,b_j)$ is the Euclidean distance between pixel $a_i$ and $b_j$.

As both edge supervision and boundary-based losses are approached from a spatial-geometric perspective, their ideological origins are consistent.
However, to the best of our knowledge, no research has emerged that have proposed an edge supervision scheme in combination with boundary-based losses, for semantic segmentation tasks.
\section{Methods}
\subsection{Edge-aware Plug-and-play Scheme}
To address the issue that the aforementioned relevant edge supervision methods are not easily applicable to other types of semantic segmentation networks, we propose an Edge-aware Plug-and-play Scheme (EPS).
It is applicable to any semantic segmentation network and is easy to use.
EPS mainly consists of two steps, which are to extract Edge GT with a thickness of $d_e$ and to copy decoder head.

Firstly, the edge thickness $d_e$ can reflect the degree of edge coarseness, and Edge GT with different edge thicknesses has different  effects on edge supervision\cite{zheng2019elkppnet}. 
In order to generate Edge GT with a thickness of $d_e$, EPS uses the simplest edge extraction method by using a kernel of size $n\times n$ to process the GT image. The relationship between $d_{e}$ and $n$, as well as the calculation method of the kernel, are as follows:
\begin{equation}
\begin{array}{rcl}
    n = 2d_e+1
\end{array}
    \label{equ:d_e}
\end{equation}
\begin{equation}
\begin{array}{rcl}\scriptsize
    kernel = 
    \begin{bmatrix}
    0 & ...&0 & -1 &0 & ... & 0 \\
    ... & ...&... & ... &... & ... & ... \\
    0 & ...&0 & -1 &0 & ... & 0 \\
    -1 & ...&-1 & 4d_e &-1& ... & -1 \\
    0 & ...&0 & -1 &0 & ... & 0 \\
    ... & ...&... & ... &... & ... & ... \\
    0 & ...&0 & -1 &0 & ... & 0 \\
    \end{bmatrix}
\end{array}
    \label{equ:kernel}
\end{equation}

Secondly, EPS copies the decoder head to generate an auxiliary head with a skip connection identical to the decoder head, but its weights are not shared with the decoder head.
 However, its supervision information is Edge GT (see Figure \ref{fig:CopyHead_only_decoder_head}).
It can be seen that EPS is an abstract strategy, and its specific implementation depends on the semantic segmentation network, so it has general applicability and is easy to implement, also plug-and-play.
EPS only participates in the updating of model parameters during training and can be completely discarded during inference, without changing the size of the model during inference. 
Although this scheme is simple, our experiments have proven its effectiveness.

\subsection{Polar Hausdorff Loss}
\begin{figure}
\centering
    \begin{minipage}[t]{0.22\textwidth}
        \centering
        \includegraphics[width=3.5cm]{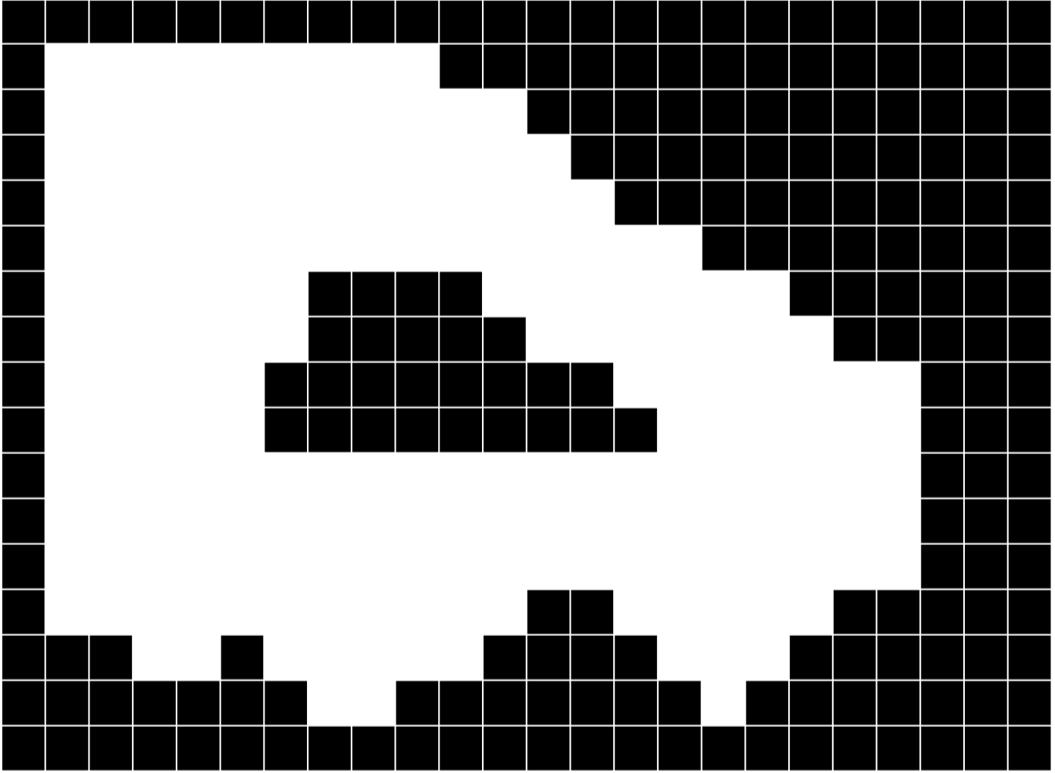}
    \end{minipage}
    \begin{minipage}[t]{0.22\textwidth}
        \centering
        \includegraphics[width=3.5cm]{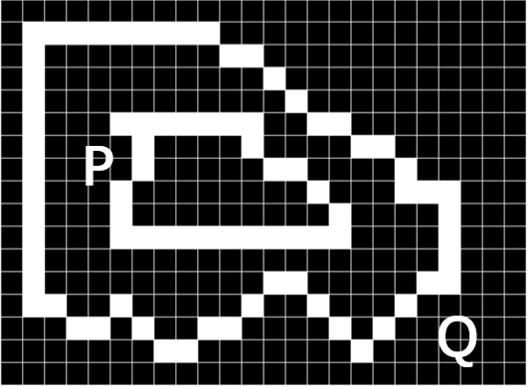}
    \end{minipage}
    \caption{On the left is the edge prediction image $\hat{p}$. On the right is the edge image $\hat{p}_e$ of $\hat{p}$, which is obtained by processing $\hat{p}$ with a thickness of $d_e=1$.}
    \label{fig:sample}
\end{figure}

In EPS, we define the edge thickness $d_{e}$ of Edge GT as prior knowledge, and its value can be reflected by the size of kernel.
For example, to set the edge thickness of Edge GT to $d_{e}=3$, a $7\times7$ kernel is used to process GT.
As the distribution-based loss optimizes globally, it doesn't have the concept of edge thickness, so the edge segmentation thickness is random and uncertain.
On the other hand, the extracted Edge GT in EPS has an explicit and equally-thick boundary with a thickness of $d_e$.
To fully utilize this prior knowledge, we propose a boundary-based loss called Polar Hausdorff (PH) Loss.
PH Loss calculates the Hausdorff distance between the internal and external edges in the predicted image $\hat{p}$ (as shown in Fig.\ref{fig:sample}), and makes it tend to $d_e$, the edge thickness of Edge GT in EPS.
This is different from HD Loss, which calculates the Hausdorff distance between the predicted image and GT.

Assuming the edge prediction image $\hat{p}$ is as shown in Fig. \ref{fig:sample} (left), the edge image $\hat{p_e}$ with thickness 1 of the predicted edge image $\hat{p}$ is extracted (Fig. \ref{fig:sample} right).
According to EPS, the edge thickness of Edge GT is defined as a preset value $d{e}$.
Therefore, during training, the Hausdorff distance between the inner edge pixel set $P$ and the outer edge pixel set $Q$ of the image in Figure \ref{fig:sample} right should tend toward $d_{e}$, which is the optimization goal of PH Loss.
Thus, PH Loss is specifically defined as:
\begin{equation}
\begin{array}{rcl}
    \mathcal L_{PH} = |PHD_{P,Q}(\hat{p},n)- d_{e}|
\end{array}
    \label{equ:L_PH}
\end{equation}
where, $PHD_{P,Q}(\hat{p},n)$ refers to the Polar Hausdorff distance between internal and external edges, and the calculation of $d_e$ is shown in equation (\ref{equ:d_e}).

 \begin{figure}[h]
    \centering
    \includegraphics[width=6.5cm]{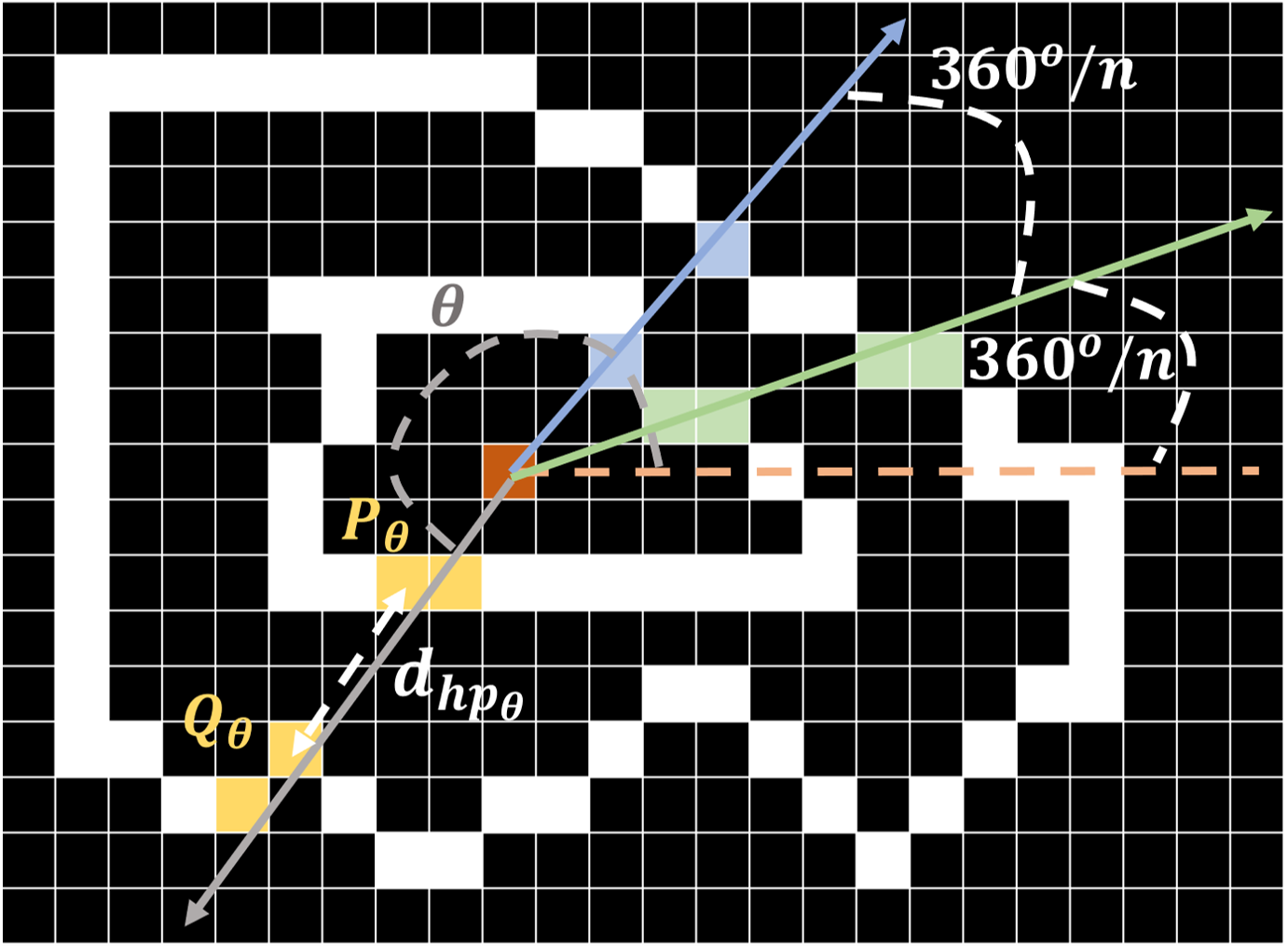}
    \caption{When computing the $PHD_{P,Q}$, the process involves drawing a ray from the polar coordinate center at an angle of $\theta=i\times 360^{\circ}/n$, and selecting the intersection points of this ray and the inner and outer edges.}
    \label{fig:polar_hausdorff}
\end{figure}

The computation of PH Loss is challenging in distinguishing the inner and outer edges of the edge image $\hat{p_e}$ for calculating the Polar Hausdorff distance between them since it is not feasible to classify them using one-pixel classifier, which can significantly increase the computational burden of the loss. 
To address this problem, we propose a more straightforward approach by utilizing the polar coordinate system to distinguish the inner and outer edges and compute their Hausdorff distance, as shown in Algorithm \ref{method:calculate_polar_hausdorff_distance}. 
Combining Fig.\ref{fig:polar_hausdorff}, we first determine the geometrical center of the set $P\cup Q$ on $\hat{p_e}$, and then use centering to convert the coordinates of all pixels in $P$ and $Q$ into polar coordinates.
We introduce a hyperparameter $n$ and draw $n$ rays from the geometric center outwards, with their angles uniformly starting from $0^{\circ}$ and increasing by $360^{\circ}/n$ each time. 
Then, for each angle $\theta$, we calculate the set of intersection points $P_{\theta}\cup Q_{\theta}$ of the ray with angle $\theta$ and the inner and outer edges, and compute the distance $\rho_{\theta}$ from each intersection point to the center. 
We select the minimum value $\rho_{min}$, and the corresponding intersection points $p$ lies on the inner edge intersection set $P_{\theta}$. 
We consider that the distances among the intersection points in $P_{\theta}$ are less than the threshold $\delta=2$, and select all the inner edge intersections $P_{\theta}$ via thresholding, while the remaining intersections $P_{\theta}\cup Q_{\theta}$ form the outer edge intersection set $Q_{\theta}$. 
Thus, we have distinguished the inner intersection sets $P_{\theta}$ and the outer intersections set $Q_{\theta}$ of the ray with angle $\theta$ between $P_{\theta}\cup Q_{\theta}$.
By traversing each angle $\theta$ of the rays, finally, we calculate the Euclidean distance $d_{ph}(\theta)$ between the farthest points in $P_{\theta}$ and the nearest points in $Q_{\theta}$.
The ultimate step is to select the maximum value from all of  $\{d_{ph}(\theta)\}$ as the Polar Hausdorff distance $PHD_{P,Q}$.

The PH Loss is a proposed edge-supervision loss combined with EPS, which involves two hyperparameters.
One is the hyperparameter for extracting the edge thickness $d_e$ of $\hat{p}_e$, which is also a hyperparameter in PH Loss.
The other is the number of rays from the polar coordinate center in PH Loss, denoted as $n$. However, in subsequent experiments, it was found that $n$ had robusness, meaning that the choice of $n$ has little effect on the results. We recommend setting $n=8$.

\section{Experiments}
 \begin{figure*}[h]
    \centering
    \includegraphics[width=17cm]{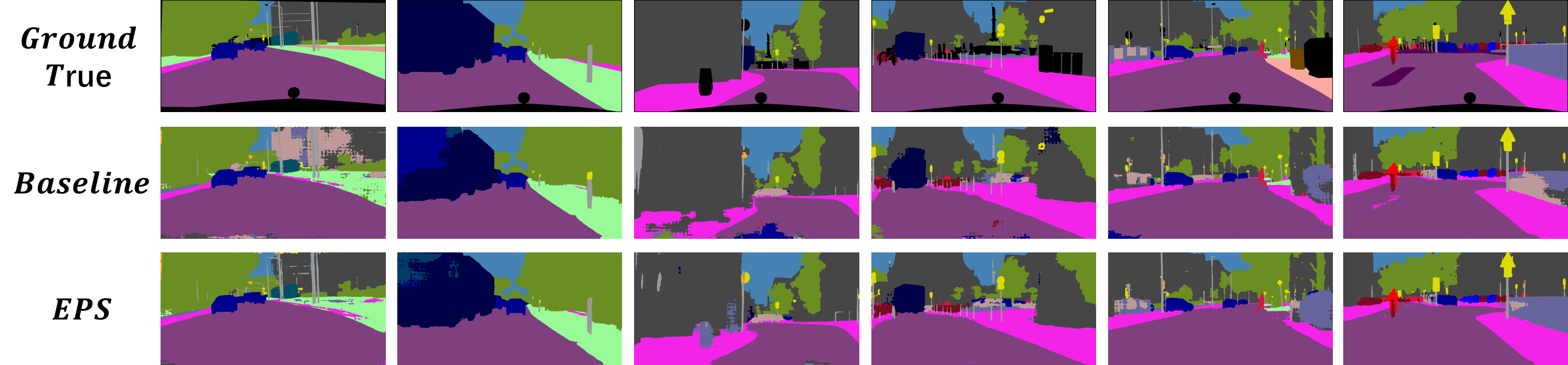}
    \caption{For the purpose of visualizing the results of semantic segmentation using CGNet, the images are presented in a top-to-bottom order, including the GT images, the baseline result, and the result obtained after applying EPS.}
    \label{fig:output}
\end{figure*}
\subsection{Experimental Settings}
Our proposed method was evaluated using the Cityscapes dataset, a well-known benchmark for semantic segmentation, which comprises 5000 high-resolution images with 19 categories. 
For hardware, we trained our models on a Linux server equipped with an I7 12700K CPU and two 24G RTX3090Ti GPUs.
In terms of software framework, we utilized the PyTorch-based semantic segmentation framework MMSegmentation for all training and testing.

To ensure fair comparisons of experimental results, all models were trained with the official default parameters of MMSegmentation, such as the learning rate and momentum. 
For the loss function, we utilized the Cross-Entropy (CE) Loss as the base and combined it with the Polar Hausdorff (PH) Loss to achieve better segmentation accuracy.
We reported segmentation accuracy using the standard mean Intersection over Union (mIoU) and mean Accuracy (mAcc) metrics.

\IncMargin{1em}
    \begin{algorithm}\small \SetKwData{Left}{left}\SetKwData{This}{this}\SetKwData{Up}{up} \SetKwFunction{Union}{Union}\SetKwFunction{FindCompress}{FindCompress} \SetKwInOut{Input}     {input}\SetKwInOut{Output}{output}
        \Input{$\hat{p}$ with size $w\times h$, $\sigma=0.1$, $\delta=2$,  $n$} 
	\Output{$PHD_{P, Q}$}
	\BlankLine 
        
        $d_{ph}=[\ ]$\;
        
	   \lIf{$\hat{p}$>0.5}
	        {$\hat{p}$ $=$ 1}
	   \lElse{$\hat{p}=$ 0}
          Get edge images $\hat{p}_{e}$ of $\hat{p}$ with $d_e=1$\;
          Get pixel index $(x,y)$ of $\hat{p}_{e}$=1\; 
          $(x_{p},y_{p})=$ $(x,y) - $mean($(x,y)$)\;
          $\rho = \sqrt{x_{p}^2+y_{p}^2}$\;
          $(cos(\alpha),sin(\alpha)) = (\frac{x_{p}}{\rho}, \frac{y_{p}}{\rho})$\;

	   \For{$j$ in arrange($n$)}
            { 
            $d_P=[\ ]$, $d_Q=[\ ]$, $d_{P\cup Q}=[\ ]$\;
	 	 \For{$(x_{i},y_{i})$ in $(x_{p},y_{p})$}
                {
                    \lIf{|$\alpha_i-j\pi/180|<\sigma$}
                    { Append $\rho_{i}$ to $d_{P\cup Q}$}
 		 	}
                $d_{min} = $Min($d_{P\cup Q}$)\;
                \For{$d_i$ in $d_{P\cup Q}$}
                {
                    \lIf{$d_i-d_{min}<\delta$}
                    { Append $d_{i}$ to $d_{P}$}
                    \lElse{Append $d_{i}$ to $d_{Q}$}
 		 	}
                Append Min($d_{Q}$) - Max($d_{P}$) to $d_{ph}$\;
 	 	}
            $PHD(P, Q) = $Max($d_{ph}$)\;
 	 	  \caption{Calculate $PHD_{P,Q}(\hat{p},n)$}
 	 	  \label{method:calculate_polar_hausdorff_distance} 
    \end{algorithm}
 \DecMargin{1em} 
 
\subsection{Performance Comparison}
\begin{table}\footnotesize
    \centering
    \setlength{\tabcolsep}{4pt}
    \caption{Comparing the results of baseline with EPS across different segmentation models.}
    \begin{tabular}{cccccc}
    \toprule  
    \multirow{2}{*}{Model} &Size& \multicolumn{2}{c}{Baseline}  & \multicolumn{2}{c}{EPS} \\
    &Params& mIoU & mAcc  & mIoU & mAcc \\
    \midrule  
    CGNet\cite{wu2020cgnet} & 496.32k & 66.84 & 80.12 & 68.68$\uparrow$ & 81.94$\uparrow$\\
    ERFNet\cite{romera2017erfnet} & 2.08M &66.08& 74.65 &70.08$\uparrow$ & 78.9$\uparrow$\\
    MobileNetV3\cite{Howard_2019_ICCV} & 3.28M & 58.12 & 67.76 & 65.82$\uparrow$ & 76.71$\uparrow$\\
    SegFormer\cite{xie2021segformer} & 3.72M & 76.28 & 83.89 & 76.88$\uparrow$ & 84.77$\uparrow$\\
    HRNet\cite{SunXLW19} & 9.64M &68.98& 77.8 & 70.09$\uparrow$ & 78.99$\uparrow$\\
    OCRNet\cite{YuanCW20} & 12.08M & 58.64 & 76.84 & 64.73$\uparrow$ & 82.54$\uparrow$\\
    ICNet\cite{zhao2018icnet} & 14.8M & 68.44 & 77.45 & 68.46$\uparrow$ & 78.27$\uparrow$\\
    STDC\cite{fan2021rethinking} & 25.17M & 53.30 & 62.55 & 58.62$\uparrow$ & 68.98$\uparrow$\\
    BiSeNetV2\cite{yu2021bisenet} & 28.5M & 64.22 & 71.19 & 65.35$\uparrow$ & 74.03$\uparrow$\\
    UNet\cite{ronneberger2015u} & 29.06 & 56.27 & 64.04 & 56.43 $\uparrow$ & 62.95\\
    PointRend\cite{kirillov2020pointrend} & 30.34M & 61.04 & 70.28 & 63.26$\uparrow$ & 71.94$\uparrow$\\
    EncNet\cite{Zhang_2018_CVPR} & 35.89M & 68.95 & 77.74 & 72.59 $\uparrow$ & 80.28 $\uparrow$\\
    EMANet\cite{li2019expectation} & 42.09M & 64.10 & 71.53 & 65.20 $\uparrow$ & 75.37 $\uparrow$\\
    ANN\cite{zhu2019asymmetric} & 46.23M & 51.26 & 63.21 & 57.17 $\uparrow$ & 66.32 $\uparrow$\\
    PSPNet\cite{zhao2017pspnet} & 48.98M & 70.32 & 78.13 & 68.51 & 76.09\\
    CCNet\cite{huang2018ccnet} & 49.83M & 60.01 & 66.88 & 60.89 $\uparrow$ & 72.78 $\uparrow$\\
    DANet\cite{fu2018dual} & 49.85M & 74.12 & 83.53 & 75.23 $\uparrow$ & 84.62 $\uparrow$\\
    NonLocal Net\cite{wang2018non} & 50.02M & 66.54 & 72.93 & 69.20 $\uparrow$ & 76.21 $\uparrow$\\
    APCNet\cite{He_2019_CVPR} & 56.36M & 45.38 & 58.45 & 51.2 $\uparrow$ & 64.72 $\uparrow$\\
    DMNet\cite{He_2019_ICCV} & 53.18M & 67.65 & 75.94 & 66.96 & 74.4 \\
    DeepLabV3\cite{chen2017rethinking} & 68.11M & 62.95 & 75.78 & 70.09 $\uparrow$ & 81.75  $\uparrow$\\
    FastFCN\cite{wu2019fastfcn} & 68.71M & 72.88 & 81.26 & 71.58 & 81.41 $\uparrow$\\
    \bottomrule 
    \end{tabular}
    \label{table:compare_EPS}
\end{table}

\begin{table}\footnotesize
    \centering
    \setlength{\tabcolsep}{2pt}
    \caption{Comparing the results of baseline with EPS $+$ PH Loss across different segmentation models.}
    \begin{tabular}{ccccccc}
    \toprule  
    \multirow{2}{*}{Model} & \multicolumn{2}{c}{Baseline}  & \multicolumn{2}{c}{EPS} & \multicolumn{2}{c}{EPS+PH}\\
    & mIoU & mAcc  & mIoU & mAcc & mIoU & mAcc \\
    \midrule  
    CGNet\cite{wu2020cgnet} & 66.84 & 80.12 & 68.68$\uparrow$ & 81.94$\uparrow$ & 70.16$\upuparrows$ & 82.26$\upuparrows$\\
    ERFNet\cite{romera2017erfnet} &66.08& 74.65 &70.08$\uparrow$ & 78.9$\uparrow$ & 71.87$\upuparrows$ & 80.84$\upuparrows$\\
    SegFormer\cite{xie2021segformer} & 76.28 & 83.89 & 76.88$\uparrow$ & 84.77$\uparrow$ & 77.03$\upuparrows$ & 85.03$\upuparrows$\\
    STDC\cite{fan2021rethinking} & 53.30 & 62.55 & 58.62$\uparrow$ & 68.98$\uparrow$ & 77.62$\upuparrows$ & 85.11$\upuparrows$\\
    
    \bottomrule 
    \end{tabular}
    \label{table:compare_EPS_PH}
\end{table}

To verify the versatility of EPS across different semantic segmentation models, we conducted experiments where our models were trained using Edge GT generated by a $5\times5$ kernel. 
We assessed nearly all semantic segmentation models within MMSegmentation, including both those with and without auxiliary heads, such as CGNet, MobileNetV3, ANN, BiSeNetV2, and others. 
The specific models utilized and corresponding experimental outcomes are presented in Table \ref{table:compare_EPS}. 
We observed a noticeable improvement in both mIoU and mAcc for almost all state-of-the-art models after implementing the EPS strategy. 
This signifies the compatibility of EPS with various semantic segmentation models and its plug-and-play nature, which enables its direct use without any consideration of the model's unique attributes.

We selected four models, namely CGNet, ERFNet, SegFormer, and STDC, to further explore the experimental results of using PH Loss in EPS.The results are presented in Table \ref{table:compare_EPS_PH}. 
Based on our analysis, it can be inferred that the utilization of EPS has resulted in substantial enhancements in the performance of the SOTA. Furthermore, the incorporation of PH Loss has demonstrated a superior improvement in the model's accuracy.

\subsection{Ablation Studies}
Our proposed PH Loss is designed to be used in conjunction with the EPS framework, with two hyperparameters: the edge thickness $d_e$ of the Edge GT and the number $n$ of candidate distances $d_{ph}$ in the $PHD_{P,Q}(\hat{p},n)$.
In order to investigate the effects of these two hyperparameters, we conducted a series of ablation experiments on CGNet, ERFNet, SegFormer, and STDC.

After conducting an analysis of Table \ref{table:ablation_de_EPS}, we observe that there is no evident regularity for selecting the hyperparameter kernel, and the optimal kernel varies among different models.
However, the performance is relatively favorable when selecting the kernel as $5\times5$ and $7\times7$.
Additionally, employing EPS led to an improvement in mIoU, irrespective of the kernel size.
Analysis of Table \ref{table:ablation_de_EPS_PH} revealed that even with PH Loss, the optimal kernel varied across different models. 
Comparing the results of the same kernel and model in Table \ref{table:ablation_de_EPS}, it was observed that almost all experiments using PH Loss performed better than those without PH Loss. 
In instances where the results of EPS with PH Loss were inferior to those of EPS with CE Loss, the cause may be attributed to insufficient training iterations.
Finally, by analyzing Table \ref{table:ablation_n_EPS_PH}, we found that the impact of selecting hyperparameter $n$ on the results is not significant, but as $n$ increases, the computational complexity of the model also increases.
Consequently, it is recommended to select a smaller value for $n$, such as $n=8$.

\begin{table}\footnotesize
    \centering
    \setlength{\tabcolsep}{1.5pt}
    \caption{Comparing the impact of hyperparameter $d_e$ on EPS.}
    \begin{tabular}{cccccccc}
    \toprule  
    \multirow{2}{*}{Method}& \multirow{2}{*}{$d_e$}& \multirow{2}{*}{kernel}& \multicolumn{4}{c}{mIoU} \\

    & & & CGNet & ERFNet &SegFormer &STDC\\
    \midrule  
    Baseline & - & - & 66.84 & 66.08 & 76.28 & 76.37 \\
    \midrule  
    \multirow{4}{*}{EPS} & 1 & 3$\times$3 & \textbf{69.18} & 66.45 & 75.38 & 76.75 \\
     & 2 & 5$\times$5 & 68.68 & \textbf{70.08} & \textbf{76.88} & 76.92  \\
     & 3 & 7$\times$7 & 68.98 & 65.51 & 75.93 & \textbf{77.48} \\
     & 5 & 11$\times$11 & 67.55 & 67.56  & 76.12  & 76.77  \\
    
    \bottomrule 
    \end{tabular}
    \label{table:ablation_de_EPS}
\end{table}

\begin{table}\footnotesize
    \centering
    \setlength{\tabcolsep}{1.5pt}
    \caption{Comparing the impact of hyperparameter $d_e$ on EPS with PH Loss ($n$=100).}
    \begin{tabular}{cccccccc}
    \toprule  
    \multirow{2}{*}{Method}& \multirow{2}{*}{$d_e$}& \multirow{2}{*}{kernel}& \multicolumn{4}{c}{mIoU} \\

    & & & CGNet & ERFNet &SegFormer &STDC\\
    \midrule  
    Baseline & - & - & 66.84 & 66.08 & 76.28 & 76.37 \\
    \midrule  
     & 1 & 3x3 & 67.74 & 70.10 & 76.58 & \textbf{77.29}  \\
    EPS & 2 & 5x5 & 68.27 & 70.71 & \textbf{77.03} & 76.81 \\
    +PH Loss & 3 & 7x7 & \textbf{70.16} & 70.43 & 76.98 & 76.77 \\
    & 5 & 11x11 & 69.05 & \textbf{71.87} & 76.90  & 76.85 \\
    \bottomrule 
    \end{tabular}
    \label{table:ablation_de_EPS_PH}
\end{table}

\begin{table}\footnotesize
    \centering
    \setlength{\tabcolsep}{1.5pt}
    \caption{Comparing the impact of hyperparameter $n$ on EPS with PH Loss ($d_e=5$).}
    \begin{tabular}{ccccccc}
    \toprule  
    \multirow{2}{*}{Method}& \multirow{2}{*}{$n$}& \multicolumn{4}{c}{mIoU} \\
    & & CGNet & ERFNet &SegFormer &STDC\\
    \midrule  
    \multirow{2}{*}{EPS}& 100  & 69.05 & 71.87 & 76.90  & 76.85 \\
    \multirow{2}{*}{+PH Loss}& 32 & 67.18 & 71.35 & 76.90 & 77.25 \\
    & 8 & 69.07 & 71.69 & 76.87 & 77.62\\
    \bottomrule 
    \end{tabular}
    \label{table:ablation_n_EPS_PH}
\end{table}

\section{Conclusion}
Our study investigated the limitations of edge supervision methods for semantic segmentation tasks, specifically the difficulty in adapting these methods to different models.
To address this challenge, we propose a novel edge supervision scheme, EPS, which duplicates the architecture decoder head for the auxiliary task with the edge supervision. By integrating the prior knowledge of edge thickness, we develop a boundary-based loss function for the thickness preserving task, which shows promising results in addressing the aforementioned challenge.
The key advantage of our EPS is its ability to seamlessly and easily integrate into any semantic segmentation model, reflecting an innovative approach to developing universally applicable strategies. 
Our experiments, conducted on 22 models using the Cityscapes dataset, demonstrate that EPS can improve upon the state-of-the-art models. 
However, we acknowledge that there are still limitations to our study of EPS, including the lack of stability and robustness analysis of PH Loss and the need for more comprehensive experiments across multiple datasets. 
In future research, we plan to explore additional Plug-and-play schemes in depth, such as incorporating texture supervision,
optimizing the calculation of PH Loss, investigating the mechanism of PH Loss;  
to extend EPS to other tasks, eg. instance segmentation and object detection; and to apply EPS in various fields, including medical imaging, autonomous driving, and industrial defect detection.


{\scriptsize
\bibliographystyle{ieee_fullname}
\bibliography{egbib}
}
\end{document}